\documentclass{article}
\usepackage{spconf,amsmath,graphicx}
\usepackage{enumitem, array, booktabs}

\usepackage[absolute]{textpos}
\textblockorigin{10mm}{10mm} 

\title{Joint Learning for Pulmonary Nodule Segmentation, Attributes and Malignancy Prediction}
\name{Botong Wu$^{1,3}$, Zhen Zhou$^{1,3}$, Jianwei Wang$^{2}$, Yizhou Wang$^{1}$}
\address{
$^1$Nat'l Engineering Laboratory for Video Technology Cooperative Medianet Innovation Center\\
Key Laboratory of Machine Perception (MoE) Sch'l of EECS, Peking University, Beijing, 100871, China \\
$^2$ National Cancer Center/Cancer Hospital, Chinese Academy of Medical Sciences and\\
Peking Union Medical College, Beijing, 100021, China \\
$^3$ Deepwise Inc., Beijing, 100085, China}
\begin{document}
%
\maketitle

\begin{textblock}{24}(2,1)
{ Accepted for publication in IEEE International Symposium on Biomedical Imaging (ISBI) 2018}
\end{textblock}

\begin{abstract}
Refer to the literature of lung nodule classification, many studies adopt Convolutional Neural Networks (CNN) to directly predict the malignancy of lung nodules with original thoracic Computed Tomography (CT) and nodule location. However, these studies cannot tell how the CNN works in terms of predicting the malignancy of the given nodule, e.g., it's hard to conclude that whether the region within the nodule or the contextual information matters according to the output of the CNN. In this paper, we propose an interpretable and multi-task learning CNN -- Joint learning for \textbf{P}ulmonary \textbf{N}odule  \textbf{S}egmentation \textbf{A}ttributes and \textbf{M}alignancy \textbf{P}rediction (PN-SAMP). It is able to not only accurately predict the malignancy of lung nodules, but also provide semantic high-level attributes as well as the areas of detected nodules. Moreover, the combination of nodule segmentation, attributes and malignancy prediction is helpful to improve the performance of each single task. In addition, inspired by the fact that radiologists often change window widths and window centers to help to make decision on uncertain nodules, PN-SAMP mixes multiple WW/WC together to gain information for the raw CT input images. To verify the effectiveness of the proposed method, the evaluation is implemented on the public LIDC-IDRI dataset, which is one of the largest dataset for lung nodule malignancy prediction. Experiments indicate that the proposed PN-SAMP achieves significant improvement with respect to lung nodule classification, and promising performance on lung nodule segmentation and attribute learning, compared with the-state-of-the-art methods.
\end{abstract}
\section{Introduction}
\label{sec:intro}
Lung cancer is one of the most malignant tumors in the world and its 5-year-survival rate is only 18\% \cite{cancer2016}. Therefore, it is important to distinguish malignancy and benign lung nodules via low-does Computed Tomography (CT) images. Naturally a fast and accurate Computer-Aid Diagnosis (CAD) system for lung cancer is urgently desired. To be practical, the CAD system should satisfy the following requirements. Firstly, it has the ability to rapidly and accurately diagnose the malignancy of pulmonary nodules, compare with radiologists. Secondly, in addition to the malignancy score, it can provide related evidences to help radiologists assess the malignancy predictions by the CAD system, such as high-level attributes (subtlety, calcification, margin, {\it etc.}) and nodule area segmentation.

To classify pulmonary nodules in terms of their malignancy and benign, traditional works which adopted different classifiers, {\it i.e.} Support Vector Machine (SVM) \cite{SVM-2015}, k-NN classifier \cite{trad2013} and Random Forests \cite{RF-2016}, with hand-crafted image features, such as HOG \cite{HOG-2005}, 3D volume feature \cite{feature-2008}, {\it etc.} achieved promising results. Due to the success of deep neural networks in natural images, recently some researchers tried to apply convolutional neural networks in pulmonary nodule detection \cite{LUNA2016,detection-2017}, segmentation \cite{seg-2016} and classification  \cite{multiscale-2015, tumornet-2017, shape-app-2016, multitask-2017}. These studies of lung nodule  classification usually extracted features via CNNs and then fed them into classifiers. However, these features are hard to understand for radiologists, which means it is not easy for them to accept the malignancy prediction without any relevant imaging findings.

To tackle the above mentioned problems, this paper proposes the PN-SAMP, which can provide rich semantic information ({\it i.e.} high-level attributes) as well as the nodule segmentation. These information will help radiologists assess the malignancy prediction produced by the CAD systems. Moreover, PN-SAMP exploits multiple window widths and window centers to enrich the nodule information. Extensive experiments show that M2LSCS achieves a significant improvement with respect to pulmonary nodule classification, promising performance for pulmonary nodule segmentation and attribute learning on LIDC-IDRI dataset  \cite{LIDC2011}. 

\begin{figure*}[htp]
	\centering
		\includegraphics[width=0.8 \textwidth] {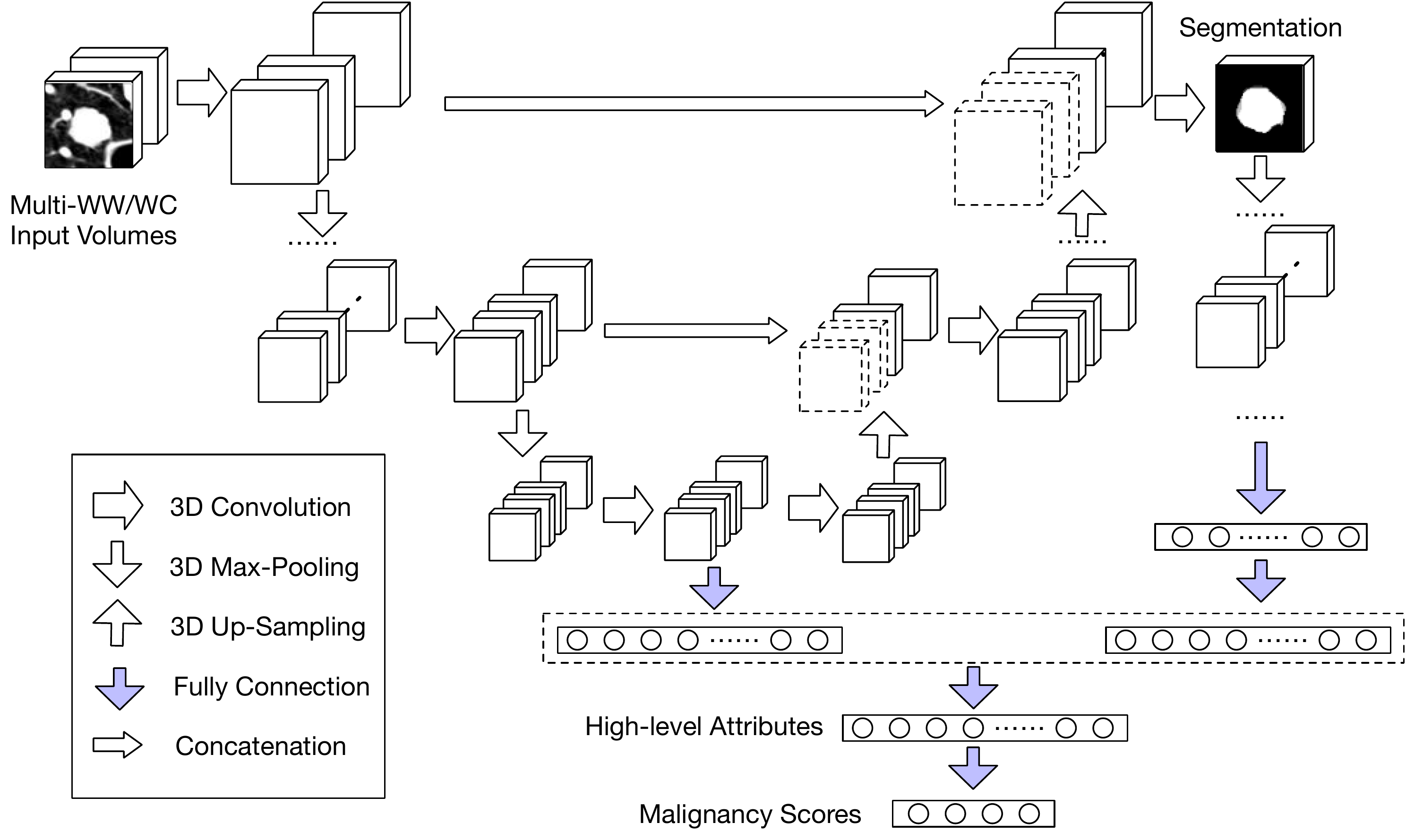}\label{subfig:wiki-map} 
	\caption{The framework of the proposed PN-SAMP. The input of the framework is a two-channel image patch containing the given nodule. The image patches of two different window widths and window centers are stacked together through the channel. Then the input are fed into the proposed neural network. Attribute learning and malignancy prediction are connected by two ways. One is from the bottom, which includes rich semantic information. The other is connected from a sub-3DCNN, the input of which is the nodule segmentation. } \label{fig:algorithm}
\end{figure*}

\subsection*{Contributions}
\begin{itemize}[leftmargin=*]
\item We propose an end-to-end multi-task and interpretable 3D convolutional neural network to simultaneously predict the malignancy of lung nodules, segment the nodule areas and learn nodule attributes. Thus, the proposed method is able to provide semantic high-level attributes as well as the region of pulmonary nodules, which makes it easy for radiologists to understand how it makes decisions. 

\item The proposed PN-SAMP utilizes multiple window widths and window centers, which enriches nodule information. Multi-task learning is also exploited to improve the performance of each task.

\item Extensive experiments are implemented on LIDC-IDRI dataset, which shows that the proposed PN-SAMP achieves the best performance in terms of segmentation, attributes and malignancy prediction, compared with previous the-state-of-the-art methods.
\end{itemize}

\section{Materials}
\label{sec:materials}
In this paper, evaluation is performed on the LIDC-IDRI dataset \cite{LIDC2011} from Lung Image Database Consortium. The LIDC-IDRI dataset includes 1010 patients (1018 scans) and 2660 nodules with slice thickness varying from 0.45 mm to 5.0 mm. There are nine labeled attributes for each nodule, {\it i.e.}, subtlety, internal structure, calcification, sphericity, margin, lobulation, spiculation, radiographic solidity and malignancy. For each nodule, its malignancy rating (attribute rating) is evaluated by radiologists, the score of which ranges from 1 to 5 where 1 denotes highly unlike malignant (highly without the given attribute) and 5 denotes highly malignant (highly with the given attribute). In our experimental setting, if mean value of malignancy ratings of the given nodule is 3, then it is ignored in experiments. The setting is also used in \cite{multitask-2017,multiscale-2015} which can avoid hindering learning process with unsure training data. The malignancy rating and the high-level attribute rating are computed as the mean value of all ratings from radiologists. Therefore, in experiments 898 benign nodules and 506 malignancy nodules are considered. A $64\times64\times64\times2$ image volume is extracted for each nodule. Then the volume will be normalized to 0.6 $mm$ along each dimension (pixel spacing and slice thickness).

\section{Methods}
\label{sec:methods}
\subsection{Multi-Task Deep Convolutional Neural Network}
With the success of UNet and its variants in pulmonary nodule segmentation \cite{UNet-2D-2015,UNet-3D-2016,V-Net-2016}, this paper exploits a similar 3D-UNet \cite{UNet-3D-2016} structure to segment nodule volumes. As indicated in Fig. \ref{fig:algorithm}, we use only half number of convolutional layers compared with 3D-UNet and the number of kernel for each convolutional layer is also reduced to half number of the 3D-UNet. The left part in Fig. \ref{fig:algorithm} includes four convolutional blocks, and each block contains a batch normalization layer \cite{batch-2015} which can normalize the distribution of the input of each convolutional block, a convolutional layer, an exponential linear units \cite{elu2015} (ELUs) activation layer which provides small activation value for nonzero values and a max-pooling layer in order. Similar to the left part, the right part consists of four convolutional blocks where max-down-pooling is replaced by up-pooling. 

As shown in Fig. \ref{fig:algorithm}, the sub-net for attribute learning and pulmonary nodule classification is divided into two parts. One is connected from the output of the left down-sampling part, which stands for high-level semantic information. The other is a sub-net which comprises of four convolutional layers using the output of the segmentation task as input. The benefit of such design is that the gradients from semantic label prediction can also propagate to the segmentation sub-net, which can improve the performance of segmentation task. This two-way connection makes it possible to update parameters from multiple tasks. Moreover, an extra fully connected layer is stacked above the concatenation of the outputs of these two sub-nets, for the sake of combine the two-way information to predict attributes. Furthermore, malignancy prediction is built upon on the feature maps of the attribute learning, which means that it takes all information into consider.

Dice coefficient loss is chosen for pulmonary nodule segmentation, which is defined as follows.
\begin{equation} \label{equ:Lw}
    L_S = 1 - \frac{2 \sum_i^N p_i g_i + \epsilon}{\sum_i^N p_i^2 + \sum_i^N g_i^2 + \epsilon},
\end{equation}
where $N$ is the number of voxels of the segmentation output. $p_i$ and $g_i$ denote the voxel of the segmented volume and ground truth ,respectively. $\epsilon$ is to avoid dividing zero. The definition of dice coefficient can be expressed as $1-L_S$. The goal is to minimize this loss. As to attribute learning and malignancy prediction, categorical cross-entropy loss is exploited. To balance classification and segmentation, a fixed trade-off parameter between classification and segmentation is applied during training.

\begin{table}[t]
\centering

\begin{tabular}{lcc}
\toprule
 \textbf{Methods} & Dice Coef. $\%$ (SEM $\%$) & Parameters  \\ \midrule
 3D-UNet \cite{UNet-3D-2016} & 71.97 (4.96) & 16M  \\
 PN-SAMP-S1 & \textbf{74.05 (3.57)} & 1.5M  \\
 PN-SAMP-S2 & 74.01(4.14) & 1.5M  \\
 PN-SAMP-M & 73.89 (3.87) & 1.5M \\ \bottomrule
 
\end{tabular}
\caption{Comparison with U-Net in terms of pulmonary nodule segmentation measured by dice coefficient and standard error of the mean(SEM). PN-SAMP-S1 means that WW/WC is 1600/-600 and PN-SAMP-S2 indicates that WW/WC is 700/-600. PN-SAMP-M represents the model using multiple window widths and window centers. }
\label{table:seg}
\end{table}

\subsection{Experiments and Results}
\label{sec:experiment}
\subsubsection{Experimental Settings}
We adopt 5 fold cross-validation over all malignant and benign nodule volumes (1404). Meanwhile, we adopt multiple window widths and window centers to preserve more information from raw CT scans. The proposed model is trained from scratch. During training, 10\% volumes are sampled as validation set. To optimize the proposed method with fast convergency speed and small error, we adopt Adam \cite{Adam-2014} optimizer and the learning rate is set to $1e-3$. In our experiment, we use the NVIDIA TITAN X pascal GPU, the test time for a single nodule patch is within 0.5 second. The overall neural network implementation used in this work is the deep learning toolkit KERAS\cite{keras-2015}.

\subsubsection{Evaluation Metric}
We adopt dice coefficient to evaluate the performance of lung nodule segmentation. We adopt ``off-by-one'' accuracy, which means that we regard an attribute/malignancy rating with $\pm 1$ as acceptable results.

\subsubsection{Evaluation on Segmentation}
Comparison between the proposed methods and 3D-UNet \cite{UNet-3D-2016} are displayed in Table \ref{table:seg}. The dice coefficients of the proposed methods with different WW/WC combination are quite close, and all proposed methods outperform 3D-UNet by an average 1.99 \%. Moreover, the proposed methods (1.5M) only using one tenth parameters compare with 3D-UNet (16M).

\begin{table}[t]
\centering

\begin{tabular}{lcc}
\toprule
 \textbf{Methods} & Accuracy $\%$ (SEM $\%$) & \# of dataset  \\ \midrule
 TumorNet \cite{tumornet-2017} & 82.47 (0.62) & 1145 \\
 TumorNet-attribute & 92.31 (1.59) & 1145 \\
 SHC-DCNN \cite{shape-app-2016} & 82.4 & 1432 \\
 MCNN \cite{multiscale-2015} & 86.84(binary) & 1100 \\
 CNN-MTL \cite{multitask-2017} & 91.26 & 1340 \\ 
 PN-SAMP-S1 &  92.03 (7.55) & 1404 \\
 PN-SAMP-S2 &  95.30 (3.99) & 1404 \\
 PN-SAMP-M & \textbf{97.58 (1.32)} & 1404 \\ \bottomrule
 
\end{tabular}
\caption{Comparison with other studies for lung nodule classification using accuracy and standard error of the mean(SEM). }
\label{table:classification}
\end{table}

\subsubsection{Evaluation on Classification}
We compare the proposed method with recent lung nodule classification works. TumorNet \cite{tumornet-2017} is a regression model and TumorNet-attribute directly uses six high-level attributes as auxiliary information. SHC-DCNN \cite{shape-app-2016} employs spherical harmonics computation and DCNN to learn shape and appearance features, then feeds these features into random forest classifier. MCNN \cite{multiscale-2015} exploits multi-scale convolutional neural network with three scale input nodule patches to learn discriminative image features, and then feeds these features into RF or SVM classifiers. CNN-MTL \cite{multitask-2017} takes each attribute classification as a task and adopts CNNs to learn a series of features for each attribute then fuses these features to predict the malignancy of pulmonary nodule. 

As shown in Table \ref{table:classification}, the proposed PN-SAMP-S1 and PN-SAMP-S2 achieve comparable or better result than existing works. The Combination of multiple window widths and window centers (PN-SAMP-M) is helpful to achieve the best performance. We also evaluate the 9 attributes (includes malignancy) with accuracy of 89.33\%.

\subsubsection{Evaluation on Different Tasks}
To verify the effectiveness of proposed multi-task way, we assess single-task model and multi-task model separately. From Table \ref{table:compare}, we can find multi-task model outperforms segmentation model and classification model with 3.69\% and 4.69\%. The explanation of the comparison results lies in two folds. On the one hand, multi-task methods contain more supervision information,  including shape and semantic information. On the other hand, the segmentation sub-net can receive the back-propagation error of classification sub-net and the classification sub-net takes the predicted segmentation as input. Thus, the proposed multi-task method is able to perform better than the single-task methods (over 3\%).

\begin{table}[]
\centering

\begin{tabular}{lcc}
\toprule
 \textbf{Methods} & Dice Coefficient $\%$ & Accuracy $\%$  \\ \midrule
 PN-SAMP-Seg & 70.23(4.19) & -  \\
 PN-SAMP-C & - & 94.87({1.03})  \\
 PN-SAMP-M & \textbf{73.89}({3.87}) & \textbf{97.58}(1.32) \\ \bottomrule
 
\end{tabular}
\caption{Comparison between multi-task and single-task proposed methods. PN-SAMP-Seg denotes the proposed method with single segmentation task and PN-SAMP-C denotes the proposed method with single classification task.}
\label{table:compare}
\end{table}

\section{Discussion and Conclusion}
\label{sec:conclusions}
In this paper, we proposed an end-to-end multi-task 3D convolutional neural networks for Pulmonary Nodule Segmentation, Attributes and Malignancy Prediction (PN-SAMP) to tackle the challenging problem on model interpretability, while previous studies only extract feature vectors and then feed them into classifiers. The proposed multi-task model can not only predict accurate malignancy rating, but also output the related information, {\it i.e.} high-level attributes and lung nodule segmentation, which can be provided to radiologists to evaluate the quality of malignancy ratings. Moreover, refer to the work habit of radiologists, the proposed method also adopted multi-WW/WC to enrich the input information from raw CT images. Extensive experiments showed that the proposed model achieved 97.58\% for nodule malignancy prediction, 89.33\% for attributes prediction and 73.89\% for nodule segmentation. 

In future work, we will focus on lung nodule malignancy prediction with pathology data and unsure labeled data. Moreover, we plan to analysis of neural networks to further increase model interpretability.

\begin{figure}[htp]
	\centering
		\includegraphics[width=0.45 \textwidth] {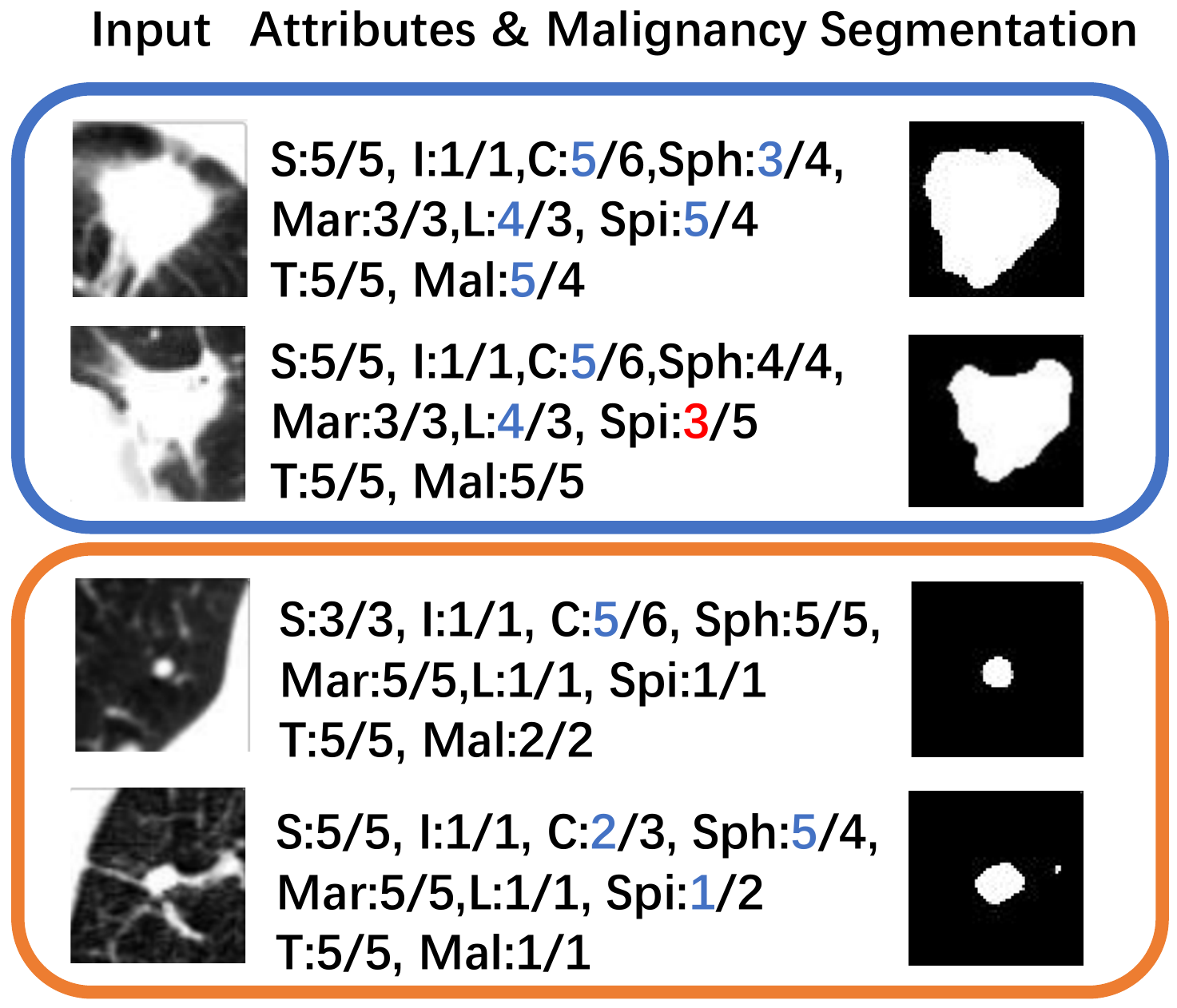} 
	\caption{Qualitative results showing attributes, malignancy rating prediction and the images on the right are segmentation results. The above part and the below part denote malignant and benign nodules. For rating A/B, A denotes the prediction of proposed method and B denotes the ground truth. Blue color integer means the prediction in range 1 of ground truth and red means the prediction out of range 1.} \label{fig:figure}
\end{figure}

\section{Acknowledgments}
This work was supported in part by 2015CB351800 and NSFC-61625201.


\bibliographystyle{IEEEbib}
\bibliography{refs}

\end{document}